\documentclass[10pt,twocolumn,letterpaper]{article}

\usepackage{cvpr}              
\usepackage{multirow}
\usepackage[accsupp]{axessibility}
\newcommand{\head}[1]{{\vspace{2.5mm}\noindent\textbf{#1}}}

\newcommand{\method}{Dream3DVG}

\usepackage{bm}

\newcommand{\R}[1]{{%
    \textbf{%
        \ifstrequal{#1}{1}{\textcolor{red}{R-hx3Y}}{%
        \ifstrequal{#1}{2}{\textcolor{blue}{R-Ztxn}}{%
        \ifstrequal{#1}{3}{\textcolor{magenta}{R-yHd7}}{%
        \ifstrequal{#1}{4}{\textcolor{teal}{R-5hfy}}{%
                           \textcolor{cyan}{R-5hfy}%
        }}}}%
    }%
}}

\newcommand{\RW}[2]{{%
    \textbf{%
        \ifstrequal{#1}{1}{\textcolor{red}{R\#hx3Y-#2}}{%
        \ifstrequal{#1}{2}{\textcolor{blue}{R\#Ztxn-#2}}{%
        \ifstrequal{#1}{3}{\textcolor{magenta}{R\#yHd7-#2}}{%
        \ifstrequal{#1}{4}{\textcolor{teal}{R\#5hfy-#2}}{%
                           \textcolor{cyan}{R\#5hfy-#2}%
        }}}}%
    }%
}}

\definecolor{cvprblue}{rgb}{0.21,0.49,0.74}
\usepackage[pagebackref,breaklinks,colorlinks,allcolors=cvprblue]{hyperref}

\def\paperID{15373} 
\def\confName{CVPR}
\def\confYear{2025}

\title{
Empowering Vector Graphics with Consistently Arbitrary Viewing and View-dependent Visibility
}

\author{
Yidi Li\textsuperscript{1}, Jun Xiao\textsuperscript{1}, Zhengda Lu\textsuperscript{1*}, Yiqun Wang\textsuperscript{2}, Haiyong Jiang\textsuperscript{1*}
\\ \textsuperscript{1} School of Artificial Intelligence, University of Chinese Academy of Sciences
\\ \textsuperscript{2} Chongqing University
\\
{\tt\small \{liyidi19@mails,xiaojun,luzhengda,haiyong.jiang\}.ucas.ac.cn, yiqun.wang@cqu.edu.cn
}
}

\begin{document}
\twocolumn[{
\maketitle
\begin{center}
    \captionsetup{type=figure}
    \includegraphics[width=\linewidth]{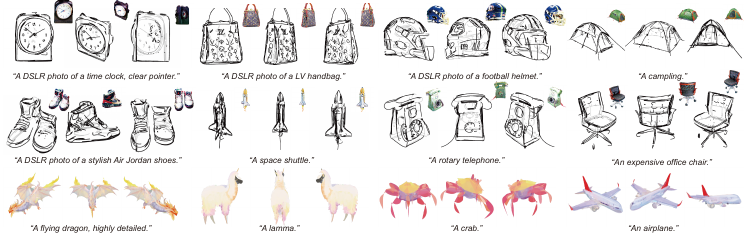}
   \caption{Examples of multiview vector graphics generated by our method conditioned on text prompts, the first two rows are sketch and the bottom row is iconography. Our method is capable of producing vector graphics with consistent views, well-preserved shape structures, and accurate occlusion relationships. The non-visible curves are rendered with lower opacity for visualization. Small images on the top right of the sketch results are the corresponding view rendering from the auxiliary 3DGS~\cite{ref/3dgs} branch as a reference for shape structures. Zoom in for details.}
   \label{fig:teaster}
\end{center}
}]

\renewcommand{\thefootnote}{\fnsymbol{footnote}}
\footnotetext{\textsuperscript{*}Joint Corresponding Authors. Haiyong is the Project Lead.} 

\begin{abstract}
This work presents a novel text-to-vector graphics generation approach, \method{}, allowing for arbitrary viewpoint viewing, progressive detail optimization, and view-dependent occlusion awareness. 
Our approach is a dual-branch optimization framework, consisting of an auxiliary 3D Gaussian Splatting optimization branch and a 3D vector graphics optimization branch. 
The introduced 3DGS branch can bridge the domain gaps between text prompts and vector graphics with more consistent guidance.
Moreover, 3DGS allows for progressive detail control by scheduling classifier-free guidance, facilitating guiding vector graphics with coarse shapes at the initial stages and finer details at later stages. 
We also improve the view-dependent occlusions by devising a visibility-awareness rendering module. 
Extensive results on 3D sketches and 3D iconographies, demonstrate the superiority of the method on different abstraction levels of details, cross-view consistency, and occlusion-aware stroke culling. 
Code is available at \href{https://github.com/chenxinl/Dream3DVG.git}{https://github.com/chenxinl/Dream3DVG.git}.



\end{abstract}    

\section{Introduction}
\label{sec:intro}


Vector graphics (VGs)~\cite{ref/svg} are widely adopted in graphic designs, art creations, and conceptual illustrations~\cite{ref/handpainter, ref/Just_DrawIt}. 
The advantage of VGs lies in their abstract styles, arbitrary resolutions, and compact sizes. 
However, VGs are designed from a specific viewpoint, and cannot support arbitrary viewing.
This hinders interesting applications, e.g., multiview sketch-based conceptual designs and cartoon animation involving different viewed VGs.  
Therefore, it is interesting to create multiview VGs with 3D consistency. 

An ideal solution for this task should satisfy the following desiderata: $\mathcal{D}1$) arbitrary viewing of VGs from different perspectives with inherent consistency; $\mathcal{D}2$) visibility-aware strokes across different views for realistic 3D perception; $\mathcal{D}3$) an accurate outline of level details in 3D shape structures; and $\mathcal{D}4$) text-driven creation for easy usage.

Recent work enables creative 2DVG generations by learning generative models for 2DVGs~\cite{ref/svg_vae, ref/deepsvg, ref/iconshop, ref/sketch_ode, ref/inrvf} and distilling the knowledge from CLIP~\cite{ref/clipasso, ref/clipascene} or pre-trained diffusion models~\cite{ref/diffsketcher, ref/vectorfusion, ref/svgdreamer}. 
Despite the wonderful results, these methods do not have 3D contexts to support arbitrary viewpoint viewing. 
On the other hand, there are very few works on 3DVG generation~\cite{ref/3doodle, ref/diff3ds}.
These works can create 3D curves for arbitrary VG viewing but their results lack a clear shape silhouette and occlusion-aware stroke culling. Furthermore, intuitively combining text-to-3D generation and image-to-VG methods~\cite{ref/clipasso} can also produce arbitrarily viewed VGs, but the cross-view consistency of strokes and clear shape structure is not guaranteed.  


In this work, we develop a novel framework for text-to-3DVG generation, called \method{}, which consists of an auxiliary 3D Gaussian Splatting (3DGS)~\cite{ref/3dgs} optimization branch and a 3D vector graphics optimization branch.
First, we utilize 3DGS as an additional branch for 3DVG optimization guidance, instead of directly optimizing 3DVGs with text prompts. 
This design offers two strengths: 1) 3DGS can ensure consistent view rendering for guidance that has smaller domain gaps than VGs to texts, and 2) A joint optimization with 3DGS allows for finer control of the 3DVG optimization. 
We progressively create varying detailed guidance images for 3DVG generation by scheduling the classifier-free guidance. The coarse-to-fine (C2F) generation enables optimizing the coarse silhouette at the initial stages and finer details at the later stages.
Therefore, the auxiliary 3DGS facilitates text-to-3DVG generation satisfying the requirements of $\mathcal{D}(1,3,4)$.

However, strokes from the occluded surface are also rendered, leading to a fake and nonrealistic look as occlusions implied by 3D curves are not considered. 
Therefore, we devise the visibility-awareness rendering module in the 3DVG optimization branch. In this module, an importance filtering strategy learns the importance of each curve at a given viewpoint and an antipodal-depth visibility voting strategy infers the occlusion relationships of each curve.

In summary, our contributions lie in the following aspects.
\begin{itemize}
    \item A joint optimization framework of 3DGS generation and 3DVG generation for text-to-3DVG generation. 
    \item A progressive optimization strategy enabling coarse-to-fine 3DVG generation with different abstraction levels of details. 
    \item A visibility-awareness rendering to remove occluded strokes for consistent shape perception. 
    \item Superior performance in both text-to-3D sketch generation and text-to-2DVG generation.
\end{itemize}





\section{Related Work}
\label{sec/related work}

\subsection{Vector Graphics Generation}
In the following, we briefly review related works on vector graphics generation. 
A typical line of work directly learns a generative model from VG datasets, 
with either explicit parametric ~\cite{ref/sketch_rnn, ref/svg_vae, ref/deepsvg, ref/deepvecfont-v2, ref/iconshop, ref/t2v_npr} or neural implicit ~\cite{ref/sketchinr, ref/sketch_ode, ref/inrvf} VG representations.
However, collecting a large VG dataset for training is tedious and the abstraction nature of stroke curves and patches further complex the problem.
Considering the duality between VG and its rendering image, differential rasterization for VG~\cite{ref/diffvg, ref/diff_compose, ref/im2vec} facilitates VG generation by optimizing the rasterization losses w.r.t. images.
Recently, the VG generation has embraced large pre-trained models, such as CLIP~\cite{ref/clip} based image vectorization~\cite{ref/clipasso, ref/clipascene} and Stable Diffusion~\cite{ref/stable_diffusion} guided text-to-vector generation, including DiffSketcher \cite{ref/diffsketcher}, VectorFusion \cite{ref/vectorfusion}, and SVGDreamer \cite{ref/svgdreamer}. 
However, due to the significant domain gap between text and vector graphics~\cite{ref/clipdraw, ref/nivel}, these methods leverage images sampled from pre-trained diffusion models as the guidance for text conditions, thereby enhancing the quality of the generated vector graphics.
These methods provide inspiration for 3DVG generation but cannot support coherent multi-view VG generation.


\subsection{Diffusion Distilled 3D Generation}
With the rise of diffusion models, Dreamfusion~\cite{ref/dreamfusion} introduces Score Distillation Sampling (SDS) to distill 3D distributions from pre-trained text-to-image diffusion models. 
This quickly motivates a series of following works~\cite{ref/sjc, ref/prolificdreamer, ref/fantasia3d, ref/magic3d, ref/magic123, ref/csd, ref/zero123, ref/added}. 
Despite promising results, SDS exhibits over-smoothing due to its feature-averaged effect~\cite{ref/luciddreamer} and over-saturation~\cite{ref/cfg} stemming from a large classifier-free guidance scale.
These shortcomings are detrimental to photorealistic rendering but may benefit some specific scenarios. 
For example, Wang \etal~\cite{ref/layered_image_vectorization_via_semantic_simplification} leverage the feature-averaged effect to smooth image details and achieve LOD image simplification. 
The simplified images serve as guidance for layered image vectorization. 
We take inspiration from the image simplification process and leverage SDS to create varying levels of pseudo images as geometric and semantic guidance for 3D vector graphics generation.

\begin{figure*}[htbp]
  \centering
   \includegraphics[width=0.9\linewidth]{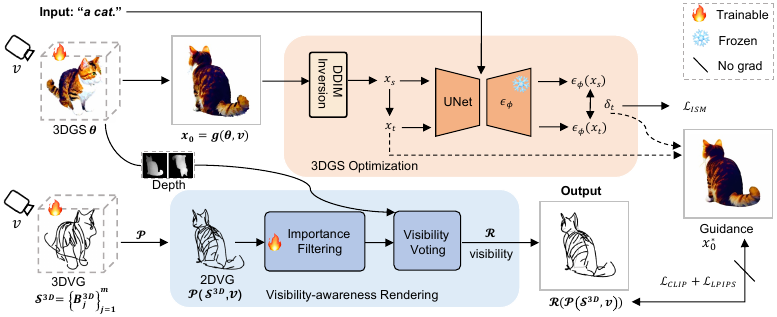}
   \caption{\textbf{The overall architecture.} The method takes as input a text prompt and outputs rendered 2D vector graphics (2DVG). The entire network consists of two branches: a 3DGS optimization branch (top row) to optimize a 3DGS with the text prompt and sample coarse-to-fine guidance images; a 3D vector graphics (3DVG) optimization branch (bottom row) that generates 3DVG and renders 2DVG with reasonable occlusion by a Visibility-awareness Rendering module.}
   \label{fig:pipeline}
\end{figure*}
\subsection{3DVG Rendering and Generation}

Compared to the wide attention paid to 2DVG generations, 3DVG generation is rarely explored. 
Some related works can generate multi-view rasterized curves from 3D meshes~\cite{ref/neural_contours} and neural radiance field~\cite{ref/ref_npr}.
However, the pixel-level output representation restricts the editability of curves. 
Recent approaches relax the limitation by introducing 3D explicit representations~\cite{ref/nef, ref/3doodle, ref/diff3ds} for vector graphics.
3Doodle~\cite{ref/3doodle} and Diff3DS~\cite{ref/diff3ds} directly generate 3D sketches as 3D B\'{e}zier curves and optimize these curves by distilling geometric and semantic knowledge from pre-trained 2D vision-language models. 
3Doodle requires multiview inputs and both two methods have some problems in achieving view-dependent curve culling and rendering.

\section{Method}
We aim at multi-view consistent 2DVG generation conditioned on a text prompt. 
Our overall framework is illustrated in Fig.~\ref{fig:pipeline}.
First, we present the 3DVG representation for multi-view 2DVG generation (Sec.~\ref{sec:preliminary}).
Then, we establish a two-branch model with the top branch for text-conditioned 3DGS generation and the lower branch for 3DVG generation with 3DGS guidance, and we jointly optimize two branches and leverage varying details of generated guidance at different diffusion steps as progressive guidance for 3DVG (Sec.~\ref{sec:ps-image}).
Finally, we improve view-dependent details of 2DVG generation by introducing visibility awareness for 3DVG rendering (Sec.~\ref{sec:render}). 

\subsection{3DVG Representation}
\label{sec:preliminary}
To ensure the consistency of multi-view 2DVG generation, we lift multi-view 2DVG in the 3D space as a set of parametric cubic B\'{e}zier curves. 
A 3DVG $\mathcal{S}^{3D}$ is organized as $n$ paths $\left\{P_i^{3D}\right\}_{i=1}^n$ with each path consisting of $m$ 3D B\'{e}zier curves $\left\{B_j^{3D}\right\}_{j=1}^m$. A 3D sketch path comprises a single curve, while a 3D iconography path consists of four curves connected end-to-end.   
Each B\'{e}zier curve $ B^{3D}$ is parameterized with four ordered 3D control points $\bm{p}^i \in \mathbb{R}^3$: 
\begin{equation}
    B^{3D}(t;\bm{p}) = \sum_{i=0}^3 b_i(t) \bm{p}^i, 0\leq t \leq 1,
\end{equation}
where $ b_i(t)\mathord{=}{3\choose i} t^i(1-t)^{3-i}$ defines the interpolation weights among four control points and ${3\choose i}$ counts the number of combinations to select $i$ points from $3$ ones.
A B\'{e}zier curve also has an associated color attribute $\bm{c}_i\in\mathbb{R}^4$ representing the RGBA color for the curve. 

The perspective projection of a 3D B\'{e}zier curve onto the image plane approximately retains its form as a 2D rational B\'{e}zier curve~\cite{ref/3doodle}, where the 2D control point $\bm{d}_{xy}^i$ are obtained by projecting the 3D control point $\bm{p}^i$ to the image plane as 2D coordinates with a perspective projection depth $\bm{d}_z^i$ from $\bm{p}^i$ to the camera center.  
We denote the perspective projected 2D B\'{e}zier curve as $\mathcal{P}(\mathcal{S}^{3D}, \bm{v})=\frac{\sum_{i=0}^3 b_i(t)\bm{d}_z^i \bm{d}_{x,y} ^i}{\sum_{i=0}^3 d_i(t)\bm{d}_z^i}$ with $\bm{v}$ denoting the projection camera pose.
Therefore, the projections of 3D sketches/iconographies maintain the mathematical formulation of 2DVGs.
The projected 2DVG can be rendered using a differentiable vector rasterizer $\mathcal{R}(\cdot)$~\cite{ref/diffvg} and optimized under the guidance of image domain, with gradient-based methods enabling end-to-end training.
See more details in Sec.~\ref{sec:vgintro} in the supplementary.





\subsection{3DVG Generation}
\label{sec:ps-image}

With a 3DVG representation, we can render multi-view 2DVGs with 3D consistency.
Our goal for view-consistent 2DVG generation can be reframed as text-conditioned 3DVG generation. 
An intuitive solution is to follow existing text-to-3D generation, e.g., DreamFusion~\cite{ref/dreamfusion}. 
However, 2DVGs, like sketches and iconographies, are underrepresented in the training dataset, leading to poor performance in distilling from a pre-trained diffusion model.
Moreover, the thin and long strip structure of rendered curves make them easily trapped into local minima, e.g., generated sketches of CLIPasso~\cite{ref/clipasso} do not fully follow the shape structure and silhouettes.
To circumvent these problems, we propose to optimize an auxiliary 3DGS denoted as $\bm{\theta}$ conditioned on a given text prompt, and utilize the \textit{optimization process} of 3DGS to guide the 3DVG generation (see Fig.~\ref{fig:pipeline}). 


 
\begin{figure}[t]
  \centering
   \includegraphics[width=0.95\linewidth]{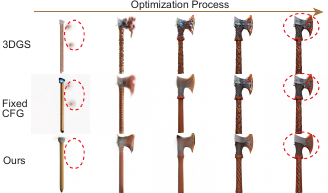}
   \caption{Guidance samples during the 3DGS optimization, with prompt \textit{``Viking axe, fantasy, weapon"}. Our sampling method (bottom row) with scheduled CFG can maintain the semantics and generate effectively smoothed samples, compared with samples from 3DGS renderings (top row) and standard sampling from diffusion trajectory with fixed CFG (second row).}
   \label{fig:simplication}
\end{figure}

In the upper row of Fig.~\ref{fig:pipeline}, the 3DGS branch takes as input a text prompt and optimizes the 3DGS $\bm{\theta}$ by distilling the pre-trained diffusion model.
The branch renders 3DGS $\bm{\theta}$ at the camera pose $\bm{v}$ as an image $\bm{x}_0$ with a differentiable renderer $\bm{g}$. 
Following ISM~\cite{ref/luciddreamer}, with different noisy latents $\bm{x}_s$ and $\bm{x}_t$ sampled through DDIM inversion~\cite{ref/ddim_inv} from $\bm{x}_0$,
we optimize 3DGS $\bm{\theta}$ with the gradient of ISM loss: 
\begin{small} 
\begin{equation}
    \nabla_\theta \mathcal{L}_{\text{ISM}} = \mathbb{E}_{t,v}
    \left[ 
        \omega(t)(\underbrace{\boldsymbol{\epsilon}_\phi(\boldsymbol{x}_t, y)-\boldsymbol{\epsilon}_\phi(\boldsymbol{x}_s)}_{\delta(t)})
        \frac{\partial \boldsymbol{g}(\bm{\theta}, \bm{v})}{\partial \bm{\theta}}
    \right],
\end{equation}
\end{small}%
where $y$ denotes the text prompt, $\boldsymbol{\epsilon}_\phi$ marks the denoiser of pre-trained diffusion models \cite{ref/ddpm, ref/sde}, $\omega(t)$ is time-dependent weights, and we denote the update direction for time step $t$ and camera pose $\bm{v}$ as $\delta(t)$.  
During the 3DGS optimization, the ISM loss searches for a deterministic trajectory in the latent space of diffusion models~\cite{ref/stable_diffusion}.
The deterministic trajectory ensures consistent semantic and geometric structure across optimization (see Fig.~\ref{fig:simplication}), avoiding noisy and corrupted updates for 3DGS generation.

Next, we optimize 3DVGs as shown in the second row of Fig.~\ref{fig:pipeline}. 
The branch produce a rasterized image of 3DVG $\mathcal{S}^{3D}$ with $\mathcal{R}(\mathcal{P}(\mathcal{S}^{3D}, \bm{v}))$ at camera pose $\bm{v}$. 
We optimize 3DVG with the guidance of 3DGS. 
Instead of directly using rendered images from 3DGS as guidance, we introduce a simple yet effective method that leverages samples from the consistent trajectory during 3DGS optimization.
Given $\boldsymbol{x}_t$ and $\delta(t)$, we obtain the guidance image $\boldsymbol{x}_0^*$ through the denoising process using a guided gradient $\delta^{*}(t)$:
\begin{equation}
    \begin{split}
        \boldsymbol{x}_0^{*} & = \frac{1}{\sqrt{\bar{\alpha}_t}}
    \left(
         \boldsymbol{x}_t - \sqrt{1 - \bar{\alpha}_t} \delta^{*}(t)
    \right), \\
    \delta^{*}(t) & = \boldsymbol{\epsilon}_\phi(\boldsymbol{x}_t)+\left[\lambda_{0} + (\lambda_{1}-\lambda_{0})(1-\frac{t}{N})\right]\delta(t),
    \end{split}
    \label{equ:cfg_interpolation}
\end{equation}
where $\bar{\alpha}_t$ is the diffusion coefficient and $N$ is the maximum timestep for the pre-trained diffusion model. 
Guided direction $\delta^{*}(t)$ is constructed based on classifier-free guidance from ISM update direction $\delta(t)$.
$\lambda_0$ and $\lambda_1$ perform guidance scale interpolation for classifier-free guidance (CFG) \cite{ref/cfg}. In this formulation, large $t$ (which corresponds to a small CFG scale) typically yields smoother $\boldsymbol{x}_0^{*}$, and as the $t$ decreases (CFG scale increases), it adds more details to $\boldsymbol{x}_0^{*}$ that align with the text prompt. This allows us to obtain samples with controllable simplification levels.
Based on this observation, we employ a time-annealing scheduler that gradually decreases the $t$-sampling range during the 3DGS optimization process, similar to~\cite{ref/dreamtime, ref/hifa}. 
Therefore, we obtain a series of $\boldsymbol{x}_0^{*}$ that become increasingly detailed throughout the 3DGS optimization process, enabling coarse-to-fine (C2F) supervision.
We show an example of different sampled guidances in Fig.~\ref{fig:simplication}.

We jointly optimize the above two branches with ISM losses for the 3DGS, the LPIPS loss~\cite{ref/lpips} and CLIP loss~\cite{ref/clipasso, ref/clipascene, ref/3doodle} for the 3DVG.
The LPIPS loss is for geometric structural constraints and the CLIP loss is for high-level semantic perception similarity:
\begin{equation}
    \begin{split}
        \mathcal{L}_{VG} = & \mathbb{E}_{\bm{v}} 
        \left[
        \mathrm{LPIPS}(\boldsymbol{x}_0^{*}, \mathcal{R}(\mathcal{P}(\mathcal{S}^{3D}, \bm{v})))\right] + \\ 
        & \mathrm{dist} \left[\mathrm{CLIP}(\boldsymbol{x}_0^{*}, \mathrm{CLIP}(\mathcal{R}(\mathcal{P}(\mathcal{S}^{3D},\boldsymbol{v})))\right],
    \end{split}
\end{equation}
where $\mathrm{dist}(\cdot,\cdot)$ denotes cosine distance, and $\mathbb{E}_{\bm{v}}$ represents the expectation over randomly sampled camera poses.

\subsection{Visibility-awareness Rendering}
\label{sec:render}
A proper rendering of $\mathcal{S}^{3D}$ must faithfully adhere to view-dependent occlusions inherent in 3D space. 
Failing to do so would introduce incorrect geometric structures in 2D renderings leading to erroneous gradients during the 3DVG optimization.
However, 3DVGs are sparsely distributed in 3D space, and the typical z-buffering for 3D occlusion does not apply due to the lack of explicit surface. 
This work proposes a visibility-awareness rendering (VAR) module to filter out occluded curves by learning the view-dependent \textit{Importance Filtering} and \textit{Antipodal-depth Visibility Voting}.

\begin{figure}[htbp]
  \centering
   \includegraphics[width=\linewidth]{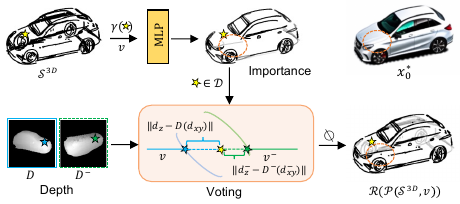}
   \caption{Illustration of Visibility-awareness Rendering. Note that we render the importance with trained opacity and non-visible curves with fixed low opacity for 2DVG rendering visualization.}
   \label{fig:rendering}
\end{figure}

\paragraph{Importance Filtering.}
The importance of a curve in 3DVG $\mathcal{S}^{3D}$ usually varies according to the viewpoint.  
For example, curves corresponding to occluded shapes or textureless regions are less important (except the silhouette under the current view), and rendering these curves with reduced opacity can enhance realism as shown in Fig.~\ref{fig:rendering}. 

With this insight, we first learn a continuous implicit function $\boldsymbol{f}$ with learnable parameter $\bm{\psi}$ to measure the importance of a 3D point $\boldsymbol{p}$ at a given view. 
We model $f$ as a multi-layer perception (MLP) neural network and the network takes as input a camera pose $\bm{v}$ and the positional encoding $\gamma(\bm{p})$ \cite{ref/nerf} of a 3D point to capture high-frequency positional information:
\begin{equation}
    f(\bm{p},\bm{v};\bm{\psi}) = \mathrm{Sigmoid}(\mathrm{MLP}(\gamma(\bm{p}), \bm{v})).
\end{equation}
Then we can treat the importance of a curve as the average importance of $k$ uniformly sampled points on the curve. 
If the importance of a curve $B^{3D}$ is lower than a pre-defined threshold $\tau_\alpha$, we think the curve can be eliminated and added to a non-important set $\mathcal{D}$.
During training, we optimize the MLP network by rendering the importance value as the opacity of a curve in $\mathcal{D}$ else a fixed high opacity and minimizing loss $\mathcal{L}_{VG}$.


\paragraph{Antipodal-depth Visibility Voting.}
Importance filtering may inevitably eliminate curves with textureless regions, causing the 3DVG to lose abstraction details and steering the optimization process toward overly simplified details. 
To maintain the integrity of the spatial structure and preserve the abstraction details, we recalibrate the visibility of filtered curves in the last step through antipodal-depth visibility voting (see Fig.~\ref{fig:rendering}).
We make the observation that a curve in $\mathcal{D}$ should be culled if it is closer to the back surface of a 3D shape than its front surface else preserved. 

For a 3D point sampled on a curve $B^{3D}\in{}\mathcal{D}$, we project the point to the front image plane viewed from camera $\bm{v}$ and the back image plane viewed from its antipodal camera $\bm{v}^{-}$. 
We denote the perspective projected 3D point in the camera space as $\mathbf{d}\in\mathbb{R}^3$ and $\mathbf{{d}^{-}}\in\mathbb{R}^3$, respectively. 
$\bm{d}$ are composed of the projected xy coordinates $\bm{d}_{xy}$ and a depth value $\bm{d}_z$.
Afterward, we render the front depths and back depths of 3DGS at two cameras $\bm{v}, \bm{v}^{-}$ as $\mathbf{D}$ and $\mathbf{{D}^{-}}$, respectively. 
We can query the depth of the underlying 3DGS shape with $\mathbf{D}(\bm{d}_{xy})$. 
Then we can determine if a curve point is closer to the front surface or the back surface through: 
\begin{equation}
    \Vert \mathbf{d}_z - \mathbf{D}(\bm{d}_{xy})\Vert - \tau_d <
    \Vert \mathbf{\bm{d}}^{-}_z - \mathbf{{D}^{-}}(\bm{d}^{-}_{xy})\Vert, 
\end{equation}
where $\tau_d =\alpha\cdot\Vert \mathbf{D}(\bm{d}_{xy})-\mathbf{{D}^{-}}(\bm{d}_{xy}^{-}) \Vert$ is an adaptive threshold depending on the thickness of a shape. 
Adding this term encourages points on the silhouette (visible to both the front and back image plane) to be visible.
We determine the visibility of a curve if the number of visible sampled points is greater than a threshold. During inference, if a curve has a predicted opacity larger than the importance threshold or satisfies depth testing, the curve is rendered with a fixed higher opacity and otherwise with a fixed lower opacity.

\section{Experiments}
We first describe our experimental details below. 

\noindent\textbf{Implementations.}
We implemented our method based on an open-source framework, PyTorch-SVGRender~\cite{ref/svgdreamer}.
3DGS is initialized with the point cloud generated by Point-E model~\cite{ref/pointe} using given text prompts, while the initial location of 3D B\'{e}zier curve control points are initialized with farthest sampled points from the point cloud.
We use Stable Diffusion 2-1-base~\cite{ref/stable_diffusion} as the pretrained diffusion model and set $\lambda_0=1.0,\lambda_1=7.5, t\in [0.1,0.9]$.
For the camera, we randomly sample the camera center with a radius from $3.5$ to $5.0$, an azimuth from $-180^\circ$ to $180^\circ$, an elevation from $45^\circ$ to $105^\circ$, and a field of view from $18^\circ$ to $36^\circ$.
We employ the Adam optimizer~\cite{ref/adam} for optimization. For 3DVG, the learning rate of control points and color parameters is set as 0.001. For 3DGS, we follow the training parameters from the original paper. We train our method for 2000 iterations with a batch size of 4, the training process takes 1 hour on an NVIDIA A100 GPU.

\begin{figure*}[htbp]
  \centering
   \includegraphics[width=0.9\linewidth]{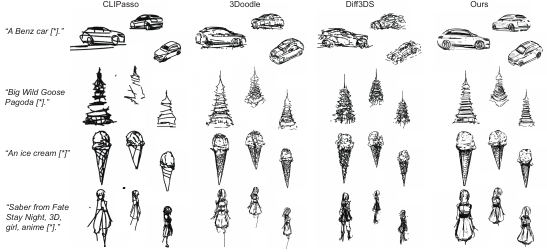}
   \caption{Qualitative results of 3D \textit{sketch}. ``\textit{[*]}" is \textit{"minimal 2d line drawing, on a white background, black and white"} for Diff3DS. All methods are rendered with the same test camera poses. The non-visible curves are rendered with lower opacity in ours.}
   \label{fig:sketch_comp}
\end{figure*}

\subsection{Comparisons}

\noindent\textbf{Evaluation Datasets and Metrics.}
We collected 40 text prompts from previous works~\cite{ref/dreamfusion, ref/luciddreamer, ref/svgdreamer} for comparisons (see details in Sec.~\ref{sec:prompts} in the supplementary). 
It is not trivial to evaluate the 3DVG generation.
In this work, we rendered the generated 3DVGs for 15 test viewpoints and adopted the metrics for 2DVG evaluation, including CLIP$^\text{text}$ and ALPIPS. 
$\mathrm{CLIP^{text}}$ compares the semantic similarity~\cite{ref/clip} between rasterized VG and the text prompt. $\mathrm{ALPIPS}$~\cite{ref/lpips, ref/tv3dg} assesses structural similarity based on the perceptual similarity between adjacent rendered views. 
Note that we use AlexNet~\cite{ref/alexnet} and Vit-B/32~\cite{ref/vit} models to calculate $\mathrm{ALPIPS}$ and $\mathrm{CLIP^{text}}$ rather than VGG16~\cite{ref/vgg} and ResNet101~\cite{ref/resnet} used in our optimization for fairness, respectively.
For $\mathrm{CLIP^{text}}$, we add the additional suffix ``\textit{minimal 2d line drawing, on a white background, black and white}" for sketch style, and ``\textit{minimal 2d vector art, linear color}" for iconography styles.

\noindent\textbf{Comparsions with Text-to-3D Sketch Generation.}
We compare our method with three typical methods, including Diff3DS~\cite{ref/diff3ds}, 3Doodle~\cite{ref/3doodle}, and a na\"{i}ve solution combining text-to-3D generation and image-to-sketch generation. 
Diff3DS uses an SDS-based method to generate 3D sketches from text with a customized differentiable rasterizer. 
Since Diff3DS has not been released, we reproduced it based on their paper with the same initialization as ours.
3Doodle~\cite{ref/3doodle} generates 3D sketches from multi-view images and does not allow for text inputs. To e
ble the comparison, we first generated 100 randomly selected viewpoints from trained text-to-3DGS generation. 
Additionally, we use the image-to-sketch method CLIPasso~\cite{ref/clipasso} to vectorize 3DGS renderings of the same test viewpoints produced by our 3DGS branch for comparisons. 
Regarding the number of paths, we use 128 for Diff3DS and 3Doodle. 
Experiments show that CLIPasso generates messy results for more than 64 strokes and we use 64 strokes for good results of CLIPasso (see Sec.~\ref{subsec:clipasso} in the supplementary).
All experiments are conducted with the same random seed. 
\begingroup
\setlength{\tabcolsep}{5pt} 
\begin{table}[]
\centering
\caption{Quantitative results for sketch and Iconography (denoted Icon here). We mark the best results in bold.}
\label{tab:comparison}
\scalebox{0.9}{
\begin{tabular}{ccccc}
\toprule
Style                        & & Method & $\mathrm{CLIP^{text}}\uparrow$ & $\mathrm{ALPIPS}\downarrow$ \\ \hline   
\multirow{7}{*}{Sketch}      & \multicolumn{1}{l}{\multirow{3}{*}{2D}} & DiffSketcher &  0.6467    &   $N/A$   \\
                             & \multicolumn{1}{l}{}                    & VectorFusion &  0.6535    &    $N/A$    \\
                             & \multicolumn{1}{l}{}                    & SVGDreamer   &  0.6574   &    $N/A$      
                             \\ \cline{2-5} 
                             & \multicolumn{1}{l}{\multirow{4}{*}{3D}}                    & CLIPasso   &  0.6602   & 0.4114\\
                             & \multicolumn{1}{l}{} & 3Doodle      &  0.6618    &   0.1718    \\
                             & \multicolumn{1}{l}{}                    & Diff3DS     &  0.6591    &   0.2046    \\
                             & \multicolumn{1}{l}{}                    & Ours         &  \textbf{0.6705}    &   \textbf{0.1656}    \\ \hline
\multirow{3}{*}{Icon} & \multicolumn{1}{l}{\multirow{2}{*}{2D}} & VectorFusion &  0.6683    &   $N/A$    \\
                             & \multicolumn{1}{l}{}                    & SVGDreamer   &   0.6740   &  $N/A$      \\ \cline{2-5} 
                             & \multicolumn{1}{l}{3D}                  & Ours         &  \textbf{0.6788}    &    0.1844  \\ 
\bottomrule
\end{tabular}}
\end{table}

Tab.~\ref{tab:comparison} shows the quantitative results, our method outperforms across both $\mathrm{CLIP^{text}}$  and $\mathrm{ALPIPS}$ for the sketch generation. 
This indicates that our method demonstrates superior performance in generating multi-view consistent 2DVG renderings and achieves the highest text-image alignment. 
We've observed that CLIPasso achieves good results in semantic similarity but underperforms in multi-view consistency, which underscores the necessity of employing 3DVG as an overall geometric representation. 

Fig.~\ref{fig:sketch_comp} shows the qualitative results. 
Diff3DS generates unorganized curves, making the results look noisy. 
Moreover, Diff3DS may struggle to generate the correct view, e.g., the saber in the front view of Fig.~\ref{fig:sketch_comp}. 
3Doodle utilizes the same 3DGS as our method, and therefore, it shares similar geometric structures as our approach with less noisy curves. 
This is because the progressive sampling strategy yields smooth and coarse-to-fine multi-view images, allowing 3Doodle to achieve more refined sketches.
Both Diff3DS and 3Doodle lack proper occlusion handling of 3D curves and still render those occluded strokes. 
CLIPasso-based 3DVG generation does not have consistent strokes across views.
Our method effectively generates multi-view consistent 3D vector graphics and accurately generates 2D renderings with accurate occlusion relationships.
Moreover, our method produces better overall shape structures, e.g., the outline of the car, repetitive structures for Goose Pagoda, a clear distinction between ice cream and the cone, and the outline of the girl's shape in Fig.~\ref{fig:sketch_comp}.

\begin{figure}[htbp]
  \centering
   \includegraphics[width=\linewidth]{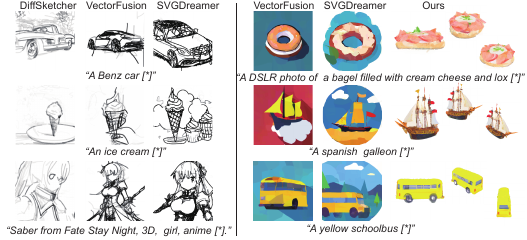}
   \caption{Qualitative results of \textit{sketch} and \textit{iconography}.}
   \label{fig:icon_comp}
\end{figure}

\noindent\textbf{Comparsions with Text-to-2DVG Generation.}
We compare against text-to-2DVG generation methods for the evaluation of single-view quality. 
There are many wonderful 2DVG generation methods and we select DiffSketcher~\cite{ref/diffsketcher}, VectorFusion~\cite{ref/vectorfusion}, and SVGDreamer~\cite{ref/svgdreamer} for comparisons.
We use their implementation and compare the method with them for the quality of a single-view rendering.

Tab.~\ref{tab:comparison} shows the results on 2D sketch generation and icon generation. 
We achieve superior performance to DiffSketcher, VectorFusion, and SVGDreamer on $\mathrm{CLIP^{text}}$ of sketches and iconography styles, suggesting our results faithfully align with the text prompts. We do not measure $\mathrm{ALPIPS}$ for 2D methods.
In Fig.~\ref{fig:icon_comp}, we show visual comparisons. 
For the sketch, DiffSketcher, VectorFusion, and SVGDreamer can produce semantically accurate VGs but may present meaningless curves (see the Benz car of SVGDreamer and the ice cream of VectorFusion in Fig.~\ref{fig:icon_comp}).
This issue is likely owing to that most training examples for pre-trained diffusion models are natural images and distilling these pre-trained diffusion models with SDS~\cite{ref/dreamfusion} for underrepresented sketches is challenging. 
For the iconography, VectorFusion and SVGDreamer yield blocky effects with uniform color (see the schoolbus in Fig.~\ref{fig:icon_comp}). 

On the contrary, our method is capable of freely expressing features of varying complexities (the bus and the bagel in Fig.~\ref{fig:icon_comp}).
This is because our iconographies, as irregular surfaces expanded in 3D space, can produce varying geometric structures when rendered from different viewpoints, enabling both simple and complex shapes to emerge and create a wider range of shape complexities.

\subsection{Ablation Studies}
In this section, we analyze the effect of important components and the hyperparameter setting of our method.

\head{3DGS Guidance.}
In Fig.~\ref{fig:ablation_module}, with initialized control points from Point-E, we start from an SDS-based~\cite{ref/dreamfusion} loss to optimize the control points, yielding intersecting and under-optimized paths, deviating from the simplicity and clarity of vector graphics (see Fig.~\ref{fig:ablation_module}-a). 
These control points, lacking geometric constraints, are prone to getting stuck in their current spatial positions and are difficult to optimize. 
Then, we substitute the SDS loss with guidance directly derived from 3DGS-renderings with $\mathcal{L}_{VG}$, bypassing SDS entirely, and the optimization of lines is enhanced with image-based supervision (see Fig.~\ref{fig:ablation_module}-b). 
Next, we employ the guidance sampled from the 3DGS optimization trajectory using a standard CFG scale of $7.5$, and the result exhibits a clear consistent shape (see Fig.~\ref{fig:ablation_module}-c). 
Finally, we dynamically adjust the CFG weights, enabling a coarse-to-fine (C2F) generation (see Fig.~\ref{fig:ablation_module}-d). 
The contours and structures of the objects become more semantically meaningful and have more reasonable details. We also show the results of using only coarse or fine guidance in Fig.~\ref{fig:coarse_fine}.

\begin{figure}[t]
  \centering
   \includegraphics[width=\linewidth]{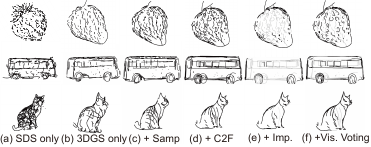}
   \caption{Visual ablations by gradually adding components.}
   \label{fig:ablation_module}
\end{figure}

\begin{figure*}[!ht]
  \centering
   \includegraphics[width=0.95\linewidth]{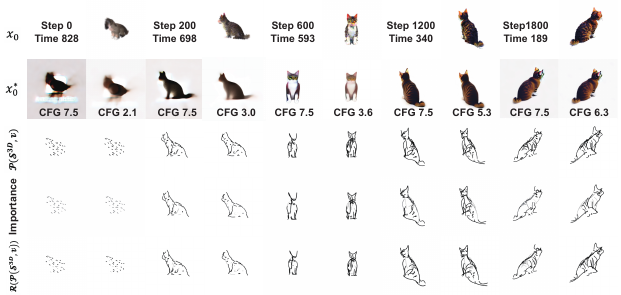}
   \caption{Guidance visualization. ``Step" is the optimization step and ``Time" is the sampled timestep $t$. We visualize the sampled image from 3DGS ($\bm{x}_0$), sampled guidance $\bm{x}_0^*$, projected 2D B\'{e}zier curves $\mathcal{P}(\mathcal{S}^{3D}, \bm{v})$ from view $\bm{v}$, the importance of curves, and the rendered image $\mathcal{R}(\mathcal{P}(\mathcal{S}^{3D}, \bm{v}))$. For each step, the first column is standard sampling, and the second column is our coarse-to-fine sampling.} 
   \vspace{-4mm}
   \label{fig:medium_visualization}
\end{figure*}
\begin{figure}[htbp]
  \centering
   \includegraphics[width=\linewidth]{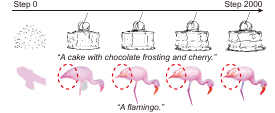}
   \caption{Illustration of the VG optimization process.}
   \label{fig:process}
\end{figure}

\head{Visibility-awareness Rendering} consists of two submodules. We first illustrate \textit{Importance Filtering} in Fig.~\ref{fig:ablation_module}-e, visualizing the unimportant curves by rendering them to a fixed lower opacity (e.g., 0.2), while a higher opacity (e.g., 1.0) for important curves. 
The contours and curves that correspond to rich semantics are deemed important, yet some details with clear semantics are underestimated. 
Then we add the \textit{Antipodal-depth Visibility Voting} module in Fig.~\ref{fig:ablation_module}-f, culling occluded curves. 
These two modules complement each other and can achieve better 3D visibility-aware rendering with important curves selected.

\head{Guidance Visualization.} 
We demonstrate the intermediate results at various optimization steps during the training process to show the effectiveness of our method in Fig.~\ref{fig:medium_visualization}. 
Compared to the images obtained from 3DGS rendering, the images we sample exhibit greater semantic smoothness, which is essential for effective supervision (see Fig.~\ref{fig:ablation_module}-b,c). Furthermore, we can see that using a constant CFG scale in sampling introduces larger gradients during the early optimization steps.
This design is detrimental to the formation of a reasonable geometric structure in the early optimization stages, while our method establishes coherent and reasonable contour lines early in the optimization process (see steps 0 and 200 in Fig.~\ref{fig:medium_visualization}). 
Therefore the results suggest that coarse-to-fine supervision is effective in generating sparse geometric structures. 
Additionally, our \textit{Visibility-awareness Rendering} can effectively infer spatial occlusion relationships with all sampling strategies.
In Fig.~\ref{fig:process}, we show our optimization process. 
Our method initially focuses on the global shape before refining local details, thereby maintaining strong semantic consistency throughout.

\begin{figure}[htbp]
  \centering
   \includegraphics[width=\linewidth]{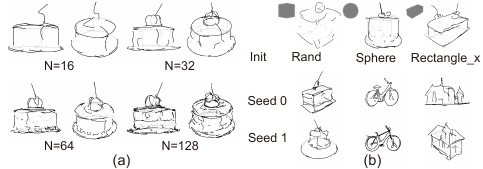}
   \caption{Abstraction and shape control. (a) The effects of using different numbers of paths $\mathrm{N}$. More paths enable finer details and less for more abstract expression. (b) The first row shows different initialization~\cite{ref/luciddreamer} and the subsequent two rows utilize different random seeds for initialization.}
   \label{fig:abstraction_level}
   \vspace{-3mm}
\end{figure}

\head{Abstraction and Shape Control.} We generate VGs with varying levels of abstraction by setting different numbers of initial paths (see Fig.~\ref{fig:abstraction_level}-a). Our method effectively captures important contours at various levels of abstraction and progressively adds details as the number of curves increases. 
Furthermore, we can control the generation of different shapes by adjusting the 3DGS initialization, such as sampling from different initial shapes or altering the random seed as shown in Fig.~\ref{fig:abstraction_level}-b. 
This flexibility provides versatility in producing a range of forms and structures, allowing for customization to meet specific design requirements or aesthetic preferences.

\begin{figure}[t]
  \centering
   \includegraphics[width=\linewidth]{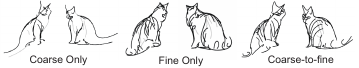}
   \caption{Effect of coarse-to-fine guidance. Relying solely on coarse guidance results in accurate geometry but losing finer details. Only fine guidance leads to incorrect geometry. A coarse-to-fine approach can balance both aspects.}
   \label{fig:coarse_fine}
   \vspace{-4mm}
\end{figure}



\section{Conclusion}

We present \method{}, a novel framework to generate arbitrary viewing vector graphics with inherent consistency and visibility awareness. 
\method{} leverages an auxiliary 3DGS generation to construct progressive guidance for 3DVG generation and introduces visibility-aware 3DVG rendering to handle view-dependent occlusions.
Extensive experiments demonstrate the effective design and the advantages of the method.

The method may produce incorrect visibility rendering (see the plate under the cake in the third row of Fig.~\ref{fig:abstraction_level}-b) when a curve spans both the current and the antipodal viewpoints since each curve is either visible or invisible.
We will explore gradient visibility and gradient colors for the curves. 
Moreover, we plan to extend this scheme to more kinds of vector graphics in future work.

\section*{Acknowledgments}
We thank all the anonymous reviewers for their insightful comments.
This work was partially supported by the National Natural Science Foundation of China (62476262, 62206263, 62271467, 62306297, 62306296, 62202076), the National Key R\&D Program of China (2024YFC3308000), Beijing Nova Program, Beijing Natural Science Foundation (4242053, L242096), and China Postdoctoral Science Foundation (2022T150639).

{
    \small
    \bibliographystyle{ieeenat_fullname}
    \bibliography{main}
}

\clearpage
\setcounter{page}{1}
\setcounter{section}{0}
\appendix
\maketitlesupplementary
\renewcommand\thesection{\Alph{section}}

\noindent In the supplementary material, we first provide illustrations for the support of different 3DVG styles and a proof of 3DVG projection (see Sec.~\ref{sec:vgintro}). Next, we show more detailed experiments in Sec.~\ref{sec:supp_ablation}. We also show how SDS influences the 3DVG results in Sec.~\ref{sec:why_not_sds}. Finally, we provide the prompts we used for comparison results in Sec.~\ref{sec:prompts}.

\section{3DVG}
\label{sec:vgintro}
\subsection{3DVG Styles}
This work shows the results for 3D sketches and 3D icons. 
For the \textit{sketch} style, each path consists of a single 3D cubic B\'{e}zier curve, with trainable control points, as illustrated in Fig.~\ref{supp_fig:curve}-a.
For the \textit{iconography} style, each path consists of four cubic 3D B\'{e}zier curves connected end-to-end to create a closed surface~\cite{ref/coons_patch}, with trainable control points and a fill surface color, as illustrated in Fig.~\ref{supp_fig:curve}-b.
Since the four curves of a 3D iconography path can construct an irregular surface expanded in 3D space, rendering it in different viewpoints may yield different 2D iconographies. 
Thus, the 3D iconography can produce different projected geometric structures from various viewpoints.

\begin{figure}[hbp]
  \centering
   \includegraphics[width=\linewidth]{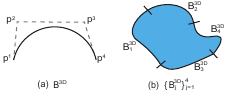}
   \caption{Illustration of two 3DVG styles, including (a) the \textit{sketch} style with a single 3D cubic B\'{e}zier curve and (b) the \textit{iconography} style with four 3D cubic B\'{e}zier curves.}
   \label{supp_fig:curve}
\end{figure}

\subsection{3DVG Projections}
Following~\cite{ref/3doodle}, we perform a perspective projection of each 3D curve. We denote the perspective projection of a 3D cubic B\'{e}zier curve $B^{3D}=B^{3D}_{xyz}=\sum_{i=0}^3{b_i(t)\bm{p}^i}$ at a camera pose $\bm{v}$ as $\mathcal{P}(B^{3D}, \bm{v})$. $\mathcal{P}(B^{3D}, \bm{v})$ is approximately identical to a 2D cubic B\'{e}zier curve $B^{2D} $ defined by $\bm{d}_{xy}=(\bm{d}^i_{xy})_{i=0,1,2,3}$, where $\bm{d}^i_{xy}$ is a perspective projection of $\bm{p}^i$.

We assume the image plane is $z$=$f$ ($f$ is the focal length) and the camera is looking at the positive $z$ direction. The perspectively projected control points for 2D B\'{e}zier curve $\mathcal{P}(B^{3D},\bm{v})$ on the image plane can be formulated as:
\begin{equation}
    \begin{split}
        & \mathcal{P}(B^{3D}, \bm{v}) =\left(
        B^{3D}_{x}\frac{f}{B^{3D}_z(t)}, B^{3D}_{y}\frac{f}{B^{3D}_z(t)}
        \right) \\
        &= \left(
        \frac{\sum_{i=0}^3b_i(t)f\bm{d}_x^i}{\sum_{i=0}^3b_i(t)\bm{d}_z^i},
        \frac{\sum_{i=0}^3b_i(t)f\bm{d}_y^i}{\sum_{i=0}^3b_i(t)\bm{d}_z^i}
        \right) \\
        &= \left(
        \frac{\sum_{i=0}^3b_i(t)\frac{f}{\bm{d}_z^i}\bm{d}_z^i\bm{d}_x^i}{\sum_{i=0}^3b_i(t)\bm{d}_z^i},
        \frac{\sum_{i=0}^3b_i(t)\frac{f}{\bm{d}_z^i}\bm{d}_z^i\bm{d}_y^i}{\sum_{i=0}^3b_i(t)\bm{d}_z^i}
        \right) \\
        &\doteq \left(
        \frac{\sum_{i=0}^3b_i(t)\bm{d}_x^i\bm{d}^i_z}{\sum_{i=0}^3b_i(t)\bm{d}_z^i},
        \frac{\sum_{i=0}^3b_i(t)\bm{d}_y^i\bm{d}^i_z}{\sum_{i=0}^3b_i(t)\bm{d}_z^i}
        \right) \\
        &\approx \left(
        \sum_{i=0}^3b_i(t)\bm{d}^i_x,
        \sum_{i=0}^3b_i(t)\bm{d}^i_y
        \right),
    \end{split}
    \label{equ:3dvg_proj}
\end{equation}
where $\bm{d}_z^i$ is the perspective projection depth from $\bm{p}^i$ to the camera center. In our implementation, $\bm{d}_z^i$ is larger enough than $\bm{d}_{xy}^i$ (5 vs. 0.5), which means the interpolation weights $\bm{p}_z^i$ of four control points are nearly the same, thus the ``$\approx$" in Eq.~\ref{equ:3dvg_proj} holds valid.
To this end, we can render the 3DVG $\mathcal{S}^{3D}$ by projecting it to 2DVG $\mathcal{P}(\mathcal{S}^{3D}, \bm{v})$ and render the 2DVG using existing 2DVG differentiable rasterzier~\cite{ref/diffvg}.


\section{Additional Experiments}
\label{sec:supp_ablation}

\begin{table}[hbt]
\setlength{\tabcolsep}{4pt}
\caption{Quantitative ablations for different modules. We started with direct 3DVG training guided by the 3DGS optimization process. Then we added the ISM module and coarse-to-fine (C2F) guidance. Next, we added importance filtering (Imp.) and depth voting (Dep.).}
\label{supp_tab:ablation}
\begin{tabular}{ccccccc}
\toprule
3DGS & ISM & C2F & Imp. & Dep. & $\mathrm{CLIP^{text}}\uparrow$ & $\mathrm{ALPIPS}\downarrow$ \\ \hline
\checkmark &  &  &  &  & 0.6461 & 0.1964 \\
\checkmark & \checkmark &  &  &  & 0.6563 & 0.1774 \\
\checkmark & \checkmark & \checkmark &  &  & 0.6658 &0.1734  \\
\checkmark & \checkmark & \checkmark & \checkmark &  & 0.6670 & 0.1748 \\
\checkmark & \checkmark & \checkmark & \checkmark & \checkmark &0.6705 & 0.1656 \\ \bottomrule
\end{tabular}
\end{table}

\begin{figure*}[htb]
  \vspace{1.25cm}
  \centering
   \includegraphics[width=\linewidth]{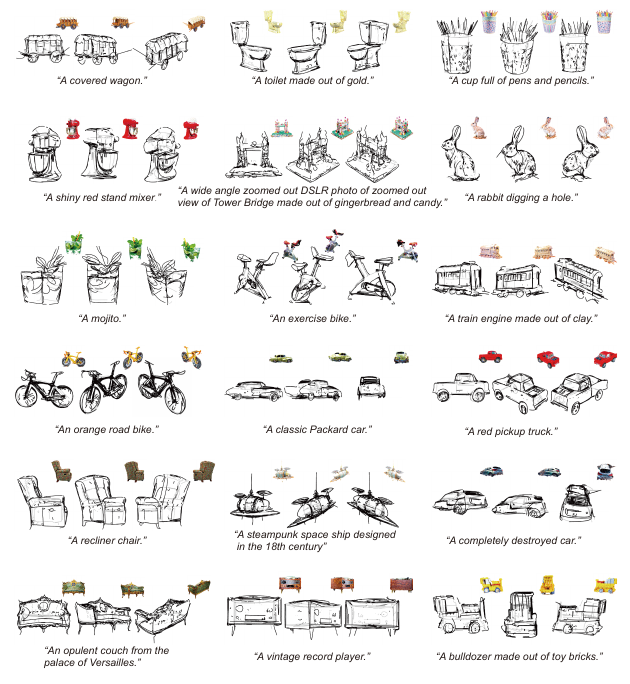}
   \caption{More examples of our text-to-3D Sketch results. Zoom in for details.}
   \label{supp_fig:sketch_results}
   \vspace{0.75cm}
\end{figure*}

\begin{figure*}[htb]
  \vspace{1.25cm}
  \centering
   \includegraphics[width=\linewidth]{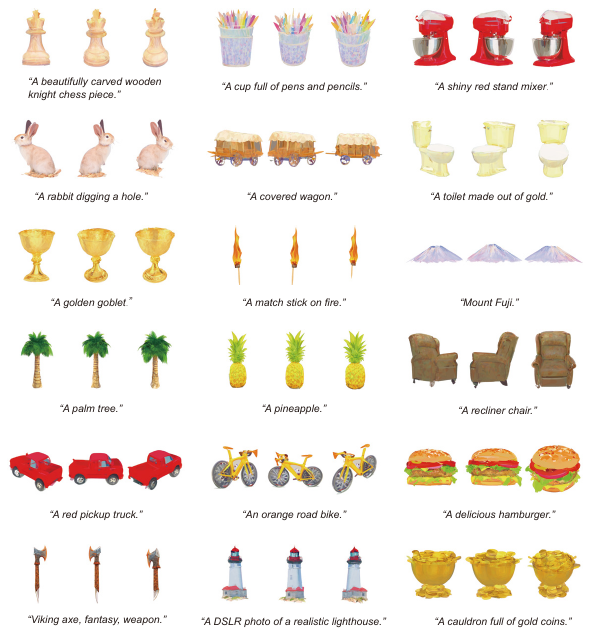}
   \caption{More examples of our text-to-3D Iconography results. Zoom in for details.}
   \label{supp_fig:icon_results}
   \vspace{0.75cm}
\end{figure*}

\begin{figure*}[htb]
  \vspace{2.5cm}
  \centering
   \includegraphics[width=\linewidth]{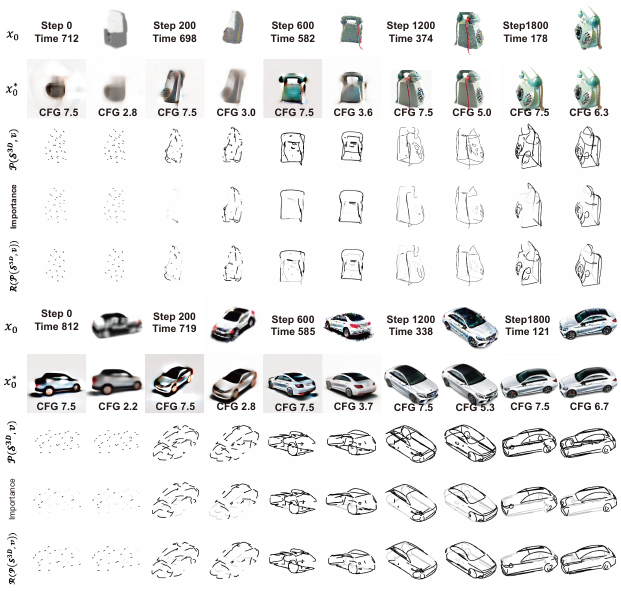}
   \caption{Guidance visualization in optimization.}
   \label{supp_fig:medium_visualization}
   \vspace{1.5cm}
\end{figure*}

\begin{figure*}[htb]
  \centering
   \includegraphics[width=\linewidth]{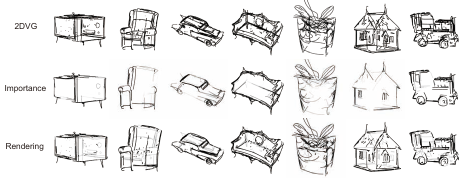}
   \caption{Visibility-awareness Rendering visualization.}
   \label{supp_fig:var_visualization}
\end{figure*}

In Tab.~\ref{supp_tab:ablation}, we ablate key components quantitatively. Results show ISM and coarse-to-fine guidance play an important role in semantic alignment (CLIP$^\text{text}$), while ISM and depth voting greatly benefit multi-view consistency (ALPIPS).
We also show more diverse results on 3D sketch generation and 3D iconography generation in Fig.~\ref{supp_fig:sketch_results} and Fig.~\ref{supp_fig:icon_results}, respectively. 
More intermediate results for key components are shown, including guidance visualization in 
Fig.~\ref{supp_fig:medium_visualization} and visibility-awareness renderings in Fig.~\ref{supp_fig:var_visualization}.
In the following, we also show validations on hyperparameters and alternative designs. 


\subsection{User study}
We present the user study results in Fig.~\ref{supp_fig:user_study}. Participants were asked to evaluate three key characteristics of 3DVG outputs: multi-view consistency, occlusion handling capability, and generation quality. Specifically, they compared multi-view renderings from four methods and selected their preferred outputs based on the following criteria:
$\mathcal{Q}1$) Which set exhibits stronger multi-view consistency?
$\mathcal{Q}2$) Which set effectively resolves spatial occlusion relationships in novel viewpoints?
$\mathcal{Q}3$) Which set better adheres to the text prompts? 

To evaluate the performance across diverse scenarios, we curated 20 text prompts and generated three representative viewpoints per prompt for each method. These multi-view visualizations were compiled into an evaluation questionnaire distributed to participants. The responses from 22 evaluators confirm the effectiveness of our approach, demonstrating superior performance in view-consistent geometry reconstruction (80\% preference rate), occlusion-aware rendering (85\% approval), and text-aligned generation quality (81\% accuracy) compared to baseline methods.

\begin{figure}[h]
  \centering
   \includegraphics[width=0.9\linewidth]{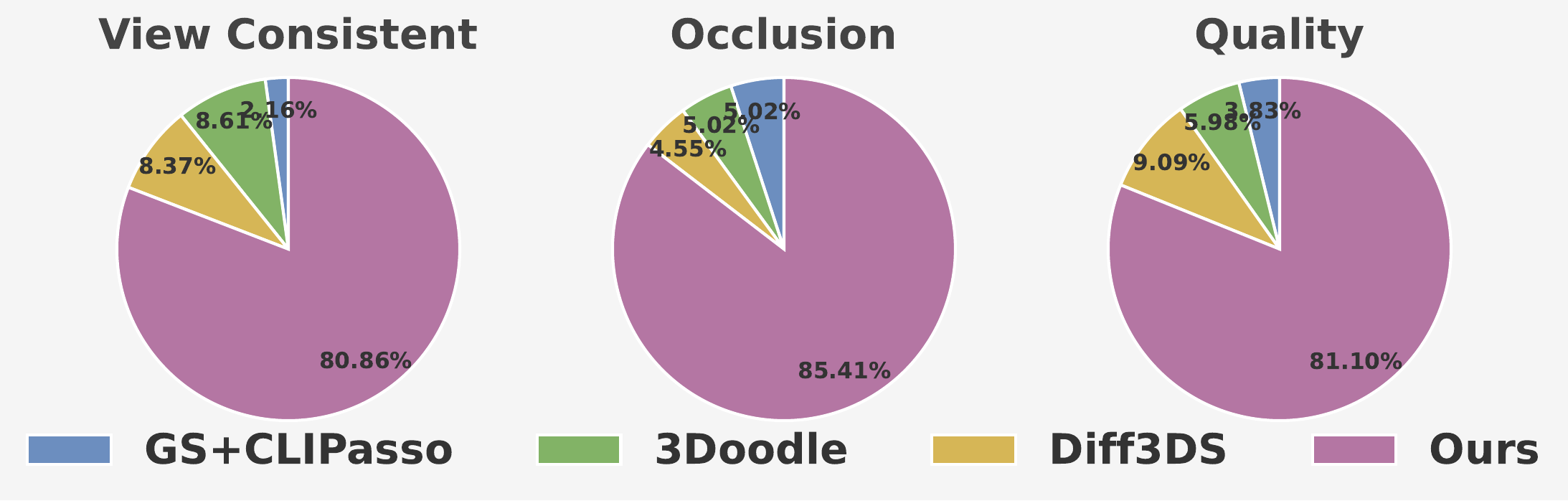}
   \caption{User study.}
   \label{supp_fig:user_study}
\end{figure}

\subsection{Sequential Optimization}
The Sequential Optimization (first the 3DGS branch then the 3DVG branch) introduces overly detailed guidance during the early stage of 3DVG optimization, leading to erroneous local curves that are challenging to correct in later stages (see Fig.~\ref{supp_fig:sequential} with zoom-in details of the guidance on the right corner). Notably, by employing a coarse-to-fine resampling strategy, the sequential approach achieves better results compared to direct optimization (3Doodle). Our joint optimization and coarse-to-fine strategy first optimize curves for a clear overall structure, then progressively add curves for details, yielding superior generation results. Additionally, the sequential optimization incurs extra diffusion sampling, leading to longer training times (see Tab.~\ref{supp_tab:train_cost}).

\begin{figure}[h]
  \centering
   \includegraphics[width=\linewidth]{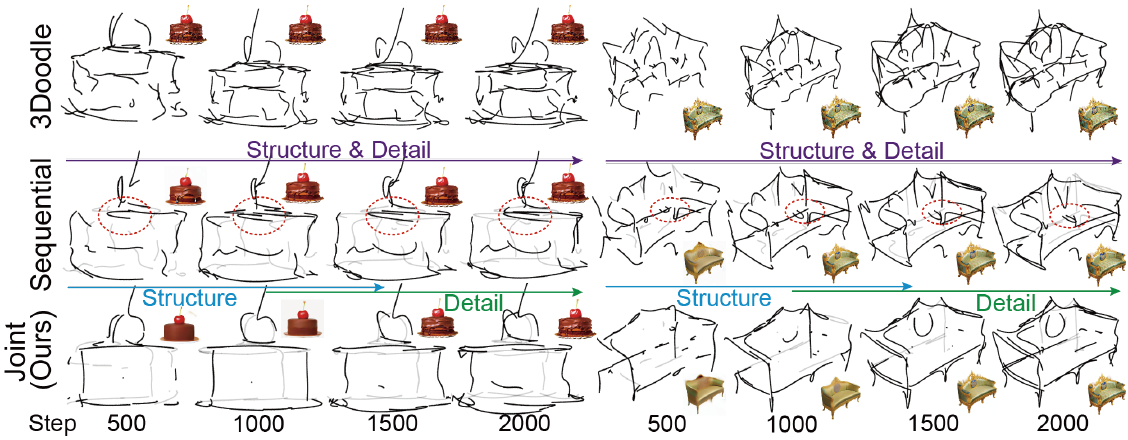}
   \caption{Sequential generation.}
   \label{supp_fig:sequential}
\end{figure}

\subsection{Training cost}
Tab.~\ref{supp_tab:train_cost} shows the training costs of different settings.

\begin{table}[hbt]
\centering
\caption{Training cost (minutes).}
\label{supp_tab:train_cost}
\scalebox{0.7}{
\begin{tabular}{cccc}
\toprule
Method & GS+CLIPasso & 3Doodle & GS+3Doodle\\ \hline
Time & 112 & 28 & 68 \\ \hline
Diff3DS & Sequential & Ours & w/o VAR \\ \hline
42 & 83 & 56 & 54\\ \bottomrule
\end{tabular}}
\end{table}

\subsection{Scene-level Generation}
Assembling our 3DVG results can produce complex scenes (see Fig.~\ref{supp_fig:scene}), and scene generation requires future work.

\begin{figure}[h]
  \centering
   \includegraphics[width=0.9\linewidth]{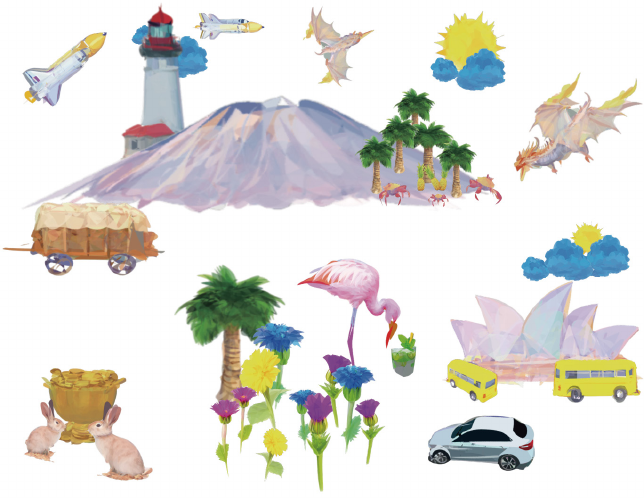}
   \caption{Scene assembling.}
   \label{supp_fig:scene}
\end{figure}

\subsection{Visibility Thresholds}
We show results on importance threshold $\tau_{\alpha}$ in Fig.~\ref{supp_fig:threshold}-a. As the threshold increases, more lines are considered to be important. We choose $\tau_{\alpha}=0.75$ to balance the number of important curves with fewer errors. 
Then we show results on depth voting threshold $\tau_{d}$ in Fig.~\ref{supp_fig:threshold}-b.

\begin{figure}[h]
  \centering
   \includegraphics[width=\linewidth]{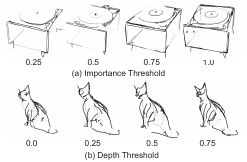}
   \caption{Ablation on visibility thresholds. (a) Importance threshold $\tau_{\alpha}$; (b) Depth voting threshold $\tau_{d}$.}
   \label{supp_fig:threshold}
\end{figure}

\subsection{The Stroke Number for CLIPasso}
\label{subsec:clipasso}
We show the influence of a large number of strokes in CLIPasso (see Fig.~\ref{supp_fig:clipasso}) to explain why we choose $N=64$ in Tab.~\ref{tab:comparison} and Fig.~\ref{fig:sketch_comp} of the main paper. A sketch with $N=32$ is too abstract to accurately represent an object, while a sketch with $N=128$ includes excessive meaningless strokes. We select the semantic best results of CLIPasso with $N=64$.
Another consideration is that only about half of the 3D sketches are visible from a specific viewpoint. 
Therefore, we need a slightly smaller number of 2D sketches than 3D sketches, ensuring relative consistency in the number of lines from corresponding viewpoints.

\begin{figure}[t]
  \centering
   \includegraphics[width=\linewidth]{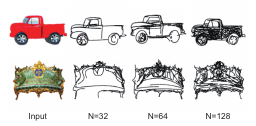}
   \caption{Results of different number of strokes in CLIPasso.}
   \label{supp_fig:clipasso}
\end{figure}

\subsection{SDS Guidance vs Our Guidance}
As shown in Fig.~\ref{supp_fig:sds_guidance}, we compare the 3DVG renderings supervised by using the guidance sampled from the SDS trajectory and our progressive guidance. 
Their guidance is derived from the same 3DGS optimization process. Our progressive guidance is capable of generating fine details. 

\begin{figure}[h]
  \centering
   \includegraphics[width=\linewidth]{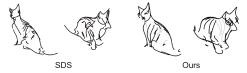}
   \caption{Ablation on the guidance type.}
   \label{supp_fig:sds_guidance}
\end{figure}





\section{Why not SDS?}
\label{sec:why_not_sds}
In this section, we first evaluate 2DVG generation results with the SDS-based guidance in Fig.~\ref{supp_fig:sds_visual}.  
We compare three optimization strategies in Fig.~\ref{supp_fig:sds_visual}-a: 1) SDS losses for 2000 steps; 2) CLIP-based loss for the initial 1000 steps and the SDS loss in the subsequent 1000 steps; and 3) the original warm-up strategy in DiffSketcher (CLIP-based loss for the initial 1000 steps, CLIP+SDS loss for the subsequent 1000 steps). 
Basically, the SDS loss generates messy curves without a clear shape structure. While the CLIP-based loss yields more promising results, the results are still disorganized by the SDS loss during the subsequent 1000 steps.

\begin{figure}[h]
  \centering
   \includegraphics[width=\linewidth]{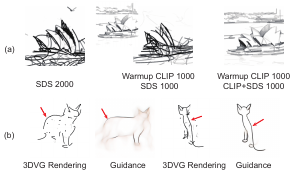}
   \caption{Effect of SDS loss.}
   \label{supp_fig:sds_visual}
\end{figure}

We further evaluate to generate 3DVGs with the SDS guidance. 
We show 3DVG rendering and the guidance images produced by SDS
with the same warmup strategy in Fig.~\ref{supp_fig:sds_visual}-b. 
We can see that the highlighted curves are not consistent between the guidance and the present 3DVG rendering. 
Therefore, gradients are not always located at the positions of the curves, making it challenging to achieve well-structured optimization. 
The SDS loss will also enhance the wrong structure as shown in the right side of Fig.~\ref{supp_fig:sds_visual}-b.

\section{Prompt for testing}
\label{sec:prompts}
\begin{itemize}
    \item[] ``A cat." 
    \item[] ``An ice cream." 
    \item[] ``A delicious hamburger."
    \item[] ``A Benz car." 
    \item[] ``A bicycle."
    \item[] ``A German shepherd."
    \item[] ``A pineapple."
    \item[] ``A ripe strawberry."
    \item[] ``A boat."
    \item[] ``A spaceship."
    \item[] ``A corgi sneezing."
    \item[] ``A pikachu."
    \item[] ``Big Wild Goose Pagoda."
    \item[] ``Sydney Opera house."
    \item[] ``Lamborghini."
    \item[] ``An airplane."
    \item[] ``A yellow schoolbus."
    \item[] ``A ceramic lion."
    \item[] ``A llama." 
    \item[] ``Flying dragon, highly detailed, breathing fire."
    \item[] ``A fire Phoenix, mythical bird, engulfed in flames."
    \item[] ``A flamingo."
    \item[] ``A Spanish galleon."
    \item[] ``A DSLR photo of a realistic lighthouse."
    \item[] ``A DSLR photo of a time clock, clear pointer."
    \item[] ``Viking axe, fantasy, weapon, blender."
    \item[] ``A DSLR photo of a bagel filled with cream cheese and lox."
    \item[] ``Saber from Fate Stay Night, 3D, girl, anime."
    \item[] ``A DSLR photo of an LV handbag."
    \item[] ``A DSLR photo of a football helmet."
    \item[] ``A DSLR photo of A Stylish Air Jordan shoes."
    \item[] ``A highly-detailed sandcastle."
    \item[] ``A yellow Swiss cheese with holes."
    \item[] ``A match stick on fire."
    \item[] ``A cake with chocolate frosting and cherry."
    \item[] ``A golden goblet."
    \item[] ``A palm tree, low poly 3d model."
    \item[] ``A Space Shuttle."
    \item[] ``A beautiful violin."
    \item[] ``A baby bunny sitting on top of a stack of pancakes."
\end{itemize}

\end{document}